\documentclass[conference]{IEEEtran}
\IEEEoverridecommandlockouts
\usepackage{cite}
\usepackage{amsmath,amssymb,amsfonts}
\usepackage{algorithmic}
\usepackage{graphicx}
\usepackage{textcomp}
\usepackage{xcolor}
\usepackage{hyperref}

\usepackage{siunitx}
\hypersetup{
    colorlinks=true,
    linkcolor=blue,
    filecolor=magenta,      
    urlcolor=blue,
    }
\def\BibTeX{{\rm B\kern-.05em{\sc i\kern-.025em b}\kern-.08em
    T\kern-.1667em\lower.7ex\hbox{E}\kern-.125emX}}

\makeatletter
\newcommand{\linebreakand}{%
  \end{@IEEEauthorhalign}
  \hfill\mbox{}\par
  \mbox{}\hfill\begin{@IEEEauthorhalign}
}
\makeatother

\begin{document}

\title{An MPSoC-based on-line Edge Infrastructure for Embedded Neuromorphic Robotic Controllers\\
\thanks{This research was partially supported by the Spanish grants MINDROB (PID2019-105556GB-C33/AEI/10.13039/501100011033), SMALL (PCI2019-111841-2/AEI/10.1309/501100011033) and COBER S.L. (Spin-off of University of Seville). E. P.-F. was supported by a "Formaci\'{o}n de Personal Universitario" Scholarship from the Spanish Ministry of Education, Culture and Sport.}
}

 \author{
 \IEEEauthorblockN{E. Piñero-Fuentes, S. Canas-Moreno, A. Rios-Navarro, D. Cascado-Caballero, A. Jimenez-Fernandez and \\ A. Linares-Barranco}
 \IEEEauthorblockA{\textit{Robotics and Technology of Computers Lab. I3US. SCORE.} \\
 \textit{University of Seville}. Seville, Spain \\
 epinerof@us.es}
}
\maketitle

\begin{abstract}

In this work, an all-in-one neuromorphic controller system with reduced latency and power consumption for a robotic arm is presented. Biological muscle movement consists of stretching and shrinking fibres via spike-commanded signals that come from motor neurons, which in turn are connected to a central pattern generator neural structure. In addition, biological systems are able to respond to diverse stimuli rather fast and efficiently, and this is based on the way information is coded within neural processes. As opposed to human-created encoding systems, neural ones use neurons and spikes to process the information and make weighted decisions based on a continuous learning process. The Event-Driven Scorbot platform (ED-Scorbot) consists of a 6 Degrees of Freedom (DoF) robotic arm whose controller implements a Spiking Proportional-Integrative-Derivative algorithm, mimicking in this way the previously commented biological systems. In this paper, we present an infrastructure upgrade to the ED-Scorbot platform, replacing the controller's hardware, which was comprised of two Spartan Field Programmable Gate Arrays (FPGAs) and a barebone computer, with an edge device, the Xilinx Zynq-7000 SoC (System on Chip) which reduces the response time, power consumption and overall complexity.

\end{abstract}


\section{Introduction}

Neuromorphic Engineering (NE) is a field in which a biological approach to day-to-day systems is searched for. Engineers try to solve traditional or new problems taking biology as a reference, just because nature's solutions are so efficient. 
Let us take a brief look at the basic processing element of nervous systems: neurons. These communicate with "spikes", instantaneous surges of electrical current that encode information in such a way that is both very efficient and completely different from the encodings engineers have created artificially. As such, NE tries to fit this "neuronal" perspective into the solutions it develops, which are as diverse as biological organisms are. Ranging in specificity and complexity, we can find works directly inspired by biological systems, like biomimetic oculomotor control\cite{biomimetic_oculomotor}, whisker control in rats\cite{whiskers} or Central-Pattern-Generation (CPG) for robots with a large number of DoF \cite{lamprea}, which can help understand how these and maybe how other, more complex systems, work. In addition to this, there is extensive research on the control topic, that is, how to make a specific system move in a controlled way so that it can achieve something meaningful. Related works using different techniques and sensing strategies include \cite{neuromorphic-control}, which uses Dynamic Adaptive Neural Network Array (DANNA) structure, \cite{pencil-control} and \cite{neuromorphic-control-dvs}, which both use Dynamic Vision Sensors (DVS) to control a certain system, \cite{SECLOC}, a control theory for Sparse Event-Based Closed Loop Control, \cite{spid}, an spiking implementation of conventional Proportional-Integrative-Derivative (PID) control and \cite{SVITE}, an spiking Vector-Integration-to-Endpoint implementation. Both latter works use Pulse-Frequency-Modulation signals to drive the motors of a robot, which is the same that we are doing here. The implementation used in this work is improved from the one used in \cite{spid}, as the feedback loop has been modified and now we're able to use it as a position controller instead of a speed controller. We now present an infrastructure upgrade to the ED-Scorbot platform\cite{EDScorbot-Springer}, designed to reduce the latency, energy consumption and overall complexity of the system, with a power consumption of 36\% of the original system and a latency improvement of 515\%.

\section{Materials and methods}

\subsection{Hardware Infrastructure}

\subsubsection{Scorbot-ER VII}
\begin{figure}[t] 
 \centering
 \includegraphics[width=8cm,height=4cm]{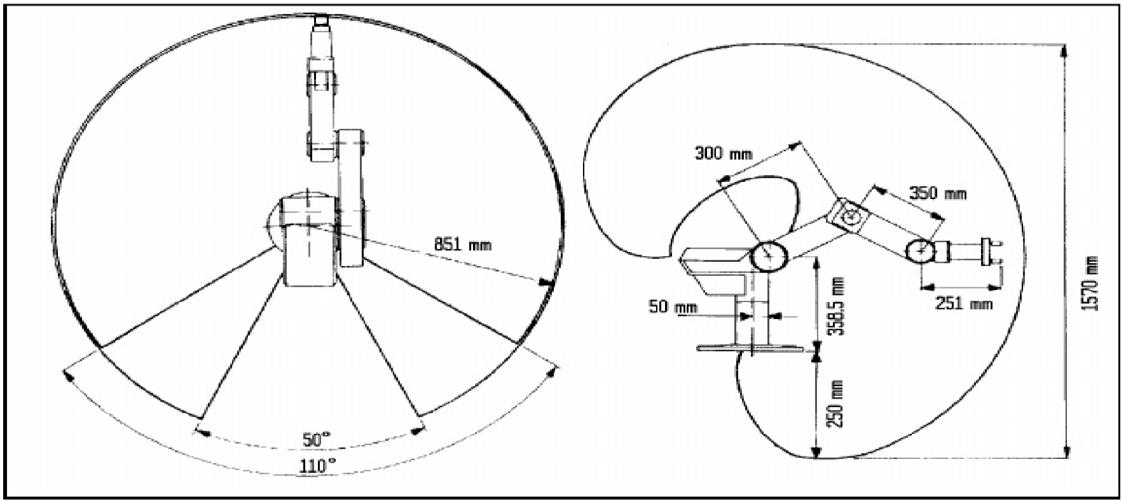}
 \caption{Scorbot ER-VII robotic diagram. Top view (left). Side view (right)}
 \label{fig:scorbot}
\end{figure}

\begin{figure}[t] 
 \centering
 \includegraphics[width=9cm]{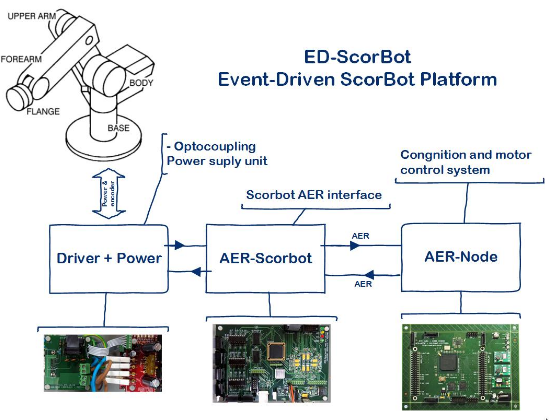}
 \caption{ED-Scorbot original infrastructure}
 \label{fig:old_platform}
\end{figure}

The robot for which the spike-based controller has been developed is a six-axis arm robot with 12V DC motors for performing axis movement and double optical encoders to register the said movements. A descriptive image of the robot can be seen in Figure \ref{fig:scorbot}. The controller developed effectively controls these six 12V DC motors through the use of a set of six H-bridges that are driven through an external 12V-25A power supply, as shown in Figure \ref{fig:old_platform}. This external supply is not affected whatsoever by the controller's programming, other than by the PFM signal that the controller generates to command different movements to each motor.

\subsubsection{Spike-based Controller}

\begin{figure}[t] 
 \centering
 \includegraphics[width=9cm]{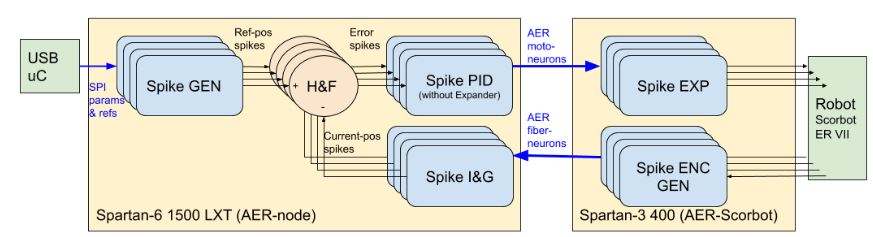}
 \caption{Block diagram of the ED-Scorbot arm spike-based PID controller for position control of the joints}
 \label{fig:SPID}
\end{figure}

The current setup for the robotic spike-based controller consists of a series of interconnected FPGAs and embedded controllers that are able to implement a communication protocol between the controller and an external device, including but not limited to a standard PC or an external spike source. This protocol consists of several different steps required to carry an input signal from a specific source to the destination, i.e., the spike-based controller. This infrastructure effectively implements an spike-based $PID$ ($SPID$) controller, modified from \cite{spid} for position control, for each joint of the robot. The architecture of the said controller is shown in Figure \ref{fig:SPID}.


The spike-based controller (1 $SPID$ per joint, 6 $SPID$ in total) is built within the AER-Node board\cite{node_board}. Configuration of $K_p$, $K_i$ and $K_d$ constants is done through the use of a series of registers, one series for each joint, as in \cite{spid}. However, due to scarcity of resources, only the $ID$ components are implemented in the AER-Node, the $P$ component is built inside the AER-Scorbot board. This second board is also necessary because it contains several optocouplers and buffers that allow for connection with DC motors as well as for feedback from the motor's encoders, the robot's home microswitches and the limit switch sensors. This feedback is of utmost importance, as the robot does not hold any state that serves as a position encoding, so it has to be implemented externally. In this implementation, a 16-bit counter per joint is used as an absolute reference for each joint's position, which is then used to close the control loop within the controller. Thus, a mapping can be created between the commanded spike input, the desired angles to reach for each joint, and its equivalent position as expressed in their counters. This mapping is shown in Table \ref{t:mappings}. Only joints one through four are accounted for, because the two other joints do not affect the end-effector Cartesian position. As it is shown, the conversion constants are different for each joint, as not all motors have the same workload nor the same operational range, which constraints all of them within a set of possible parameters. Using this mapping, a quick estimation can be made regarding the upper and lower bounds in terms of the spiking input signal and the counter assigned to each joint. Said bounds are shown in table \ref{t:bounds}. This information could be used for implementing high-level constraints, e.g., to create a Deep Learning model that generates trajectories for this particular robot. 

\begin{table}
\centering
\caption{$O$ is an offset with a value of 32768. $P$: position. $SI$:Spiking Input }
\begin{tabular}{|c|c|c|c|} 
 \hline
 Joint & SI/Degree & P/Degree & P/SI \\ 
 \hline
 J1 & \num{-3.1e-1} & \num{7.98e-3}*($P$-$O$) & \num{2.47e-2}*($P$-$O$)\\ 
 J2 & \num{-1.1e-1}  & \num{7.67e-3}*($P$-$O$) & \num{6.77e-2}*($P$-$O$) \\ 
 J3 & \num{-2.9e-1} & \num{7.05e-3}*($P$-$O$) & \num{2.39e-2}*($P$-$O$) \\ 
 J4 & \num{-5.7e-2}  & \num{1.24e-2}*($P$-$O$) & \num{2.18e-1}*($P$-$O$) \\ 
 \hline
\end{tabular}

\label{t:mappings}
\end{table}

\begin{table}
\centering
\caption{Upper and Lower 
bounds for joints one through four. 
}

\begin{tabular}{|c|c|c|c|c|} 
 \hline
 Joint & Upper(SI) & Lower(SI) & Upper (Deg)& Lower (Deg) \\ 
 \hline
 J1 &  487 & -487 & 155 & -155\\ 
 J2 &  750  & -750 & 85 & -85 \\ 
 J3 & 383 & -383 & 112.5 & 112.5 \\ 
 J4 & 1585 & -1585  & 90 & -90 \\ 
 \hline
\end{tabular}

\label{t:bounds}
\end{table}

On top of all that, the AER-Scorbot board is used as our spiking input. It lets us connect to it via USB connection (through an in-board micro-controller), which means that the controller can be commanded using an external software that handles the infrastructure configuration. 
However, it is also possible to use other sources of input for the controller, such as spiking outputs from other systems, by changing the AER-Node board's configuration. Data transfers between FPGA boards is done through an SPI bus, which turns out to be the system's bottleneck with a minimum period of 100~120ms to process a command.

Lastly, we have to talk about the actual driving of the motors or power stage. There is a minor drawback: the AER-Scorbot board is not able to drive a 12V DC motor on its own, let alone six of them. Therefore, power boards are needed for each joint, which receive the output spiking signal generated by the controller and converts it into a proportional voltage signal, strong enough to drive the motors. All mentioned hardware is shown in Figure \ref{fig:old_platform}.

With this infrastructure in mind, we intend to replace the two old Spartan boards with a single SoC comprising both an FPGA and an embedded processor so that we may remove the current setup (which includes nontrivial, slow and potentially error-prone wiring and the need of an external, dedicated PC) and just use a single board that will act both as the controller and as the external platform from which the controller is commanded. The resulting infrastructure is shown logically in Figure \ref{fig:zynq_block_diagram}. The power stage of the infrastructure is not meant to be replaced. The new SoC acting as a replacement is described in the next subsection.

\subsubsection{Zynq}
The new hardware platform is based on a Xilinx Zynq-7000® All Programmable SoC Mini-Module Plus 7Z100 (MMP). This heterogeneous platform consists mainly of two separate parts, the processing system (PS) and the programmable logic (PL). The PS has ARM cores where an embedded operating system, called Petalinux, can be installed. On the other hand, the PL has an FPGA Kintex 7 family, which is more than big enough to hold both old Spartan projects.

\begin{figure}[t] 
 \centering
 \includegraphics[width=9cm]{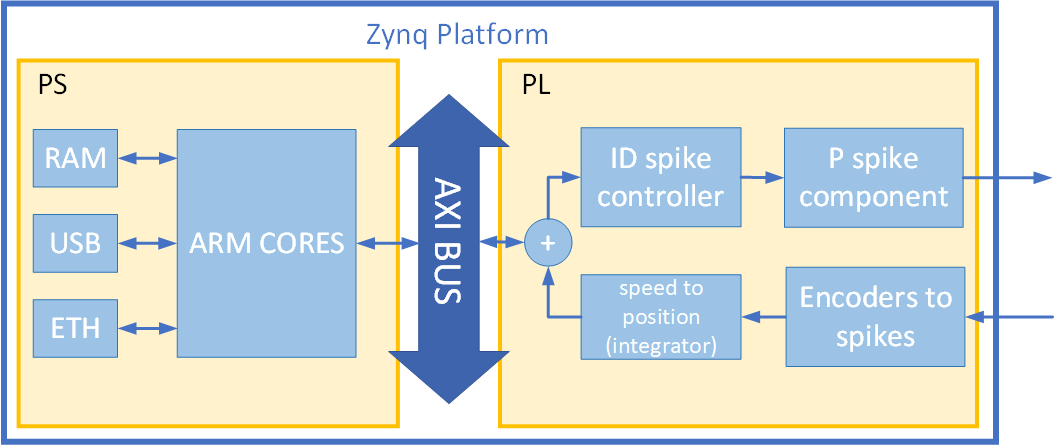}
 \caption{Hardware platform simplified block diagram}
 \label{fig:zynq_block_diagram}
\end{figure}

A simplified block diagram of the hardware platform is shown in the figure \ref{fig:zynq_block_diagram}. On the left side of the figure is the PS where the OS will run and which will have access to some peripherals such as USB or Ethernet, among others. The Ethernet interface can be used to connect to the platform via ssh, thus having an embedded system that can be easily controlled remotely. On the right side we have the PL where two IP modules have been deployed, one for each project implemented in the two old platforms. The interface of the robot is connected to the ED-Scorbot module. The connection between PS and PL is made over a standard AXI bus, which improves the performance over the SPI used in the previous setup.

\subsection{Software Infrastructure}


The software framework used for commanding the controller when it was first created is jAER, which is a commonly known piece of software for neuromorphic applications (audio/video processing, filters, etc.) that allows for, for example, monitoring and recording the output of a neuromorphic camera. As such, this was the first option that was considered when dealing with a spike-based controller. The interface developed was a purely graphical one \cite{jAER}
. With this implementation, someone with physical access to the robot could command it to perform arbitrary movements, change the controller's configuration (mainly $K_p$, $K_i$ and $K_d$ constants), etc., everything done through the graphical interface. This was very convenient at the time, as it allowed for a real-time demonstration that the spike-based controller worked, as well as hot switching and manual tuning of the SPID constants.

However, a Python software was developed mimicking the behaviour of the jAER implementation because that way the robot would be made more accessible. The newly developed tool offers some improvements, such as 100\% Python code or use through scripting, which allows for automatic execution of trajectories. It can also be used telematically, providing remote access on demand to the robot. Additionally, Python has become the go-to programming language for a big part of the Computer Science community, so we hope this provides better integration with already developed AI frameworks. 

Therefore, as of now, the preferred way to manipulate the robot is to use the new Python software, as not only does it allow for trajectory execution and data collection but it also is accessible remotely. This way, the platform can be opened up for external use. The source code of this tool, as well as further documentation on it and contact information, can be found in the tool's own repository \footnote{https://github.com/RTC-research-group/Py-EDScorbotTool, Last accessed on 18 January 2022}.
The next section will give a detail on the old Spartan functionality integration into the project's progression.
\section{Development}

The aim of this project is to reduce the latency and energy consumption derived from a heterogeneus system and old hardware. The work done consists of the two old FPGA's replacement by the new Zynq platform, having been both described previously. To do so, the two separate VHDL projects used to program the current FPGAs have been combined into one single project. This way, the same hardware structure can be maintained while also replacing both boards. The reasons for the replacement come from the various inconveniences that supporting old hardware and components present, such as the use of deprecated tools, like Xilinx's ISE Design Suite which is needed to program the Spartan-3 and Spartan-6 platforms, and from the nontrivial wirings that are needed for the system to successfully establish a communication between these two FPGAs. Furthermore, as we are using a SoC, meaning we have access to both a PL system and a PS, there is no more need for an external, dedicated PC, as one can simply use the SoC's PS as the external system that commands the controller. This will also keep the ability to have a remote connection, as the PS will offer the same features the current dedicated server does. This SoC is mounted over a carrier board called SoC-Dock.
    

The first consideration when working with this Zynq board is the clock source. In this case, we have a 200 MHz system clock instead of a 50 MHz one (Spartan). It is essential to get a 50MHz clock to avoid recalibrating all SPID controllers of the robot. 
In addition, this board has a LVDS differential clock which needs to be changed into a single-ended one. For that purpose, a buffer (IBUFDS) was used.

To be able to communicate between the PS and the PL, an AXI peripheral (AXI-4 Lite) was created, thus removing the SPI interface bus and the bottleneck of the system. The AXI interface removes a previous 8051 MCU, which was acting as a USB to SPI interface. This AXI peripheral has 36 registers of 32 bits each. The data for configuration and execution needed by the SPID controller is stored in these registers. They are mapped in memory starting from a given, user-specified address.



In the PS side, an embedded, minimal-weight, Linux-based operating system 
was installed, as it has already been commented. In this operating system, the already mentioned Python application was adapted to use the \emph{DEVMEM} driver. This adaption was developed to directly read and write to and from all registers defined by the AXI peripheral, being thus able to command the controller. This Operating System will be connected to the Internet to allow the remote access of the robot, making a transparent change for robot users connection wise.

However, this new FPGA implementation is, again, not enough to make the robotic arm work. A new PCB with six optocouplers and six 3.3V-5V level converters (one for each joint) has been designed. This PCB is necessary to make the adaptation from the previous setup to the new one, since the AER-Scorbot already integrated them. 

\subsection{Tests}

\begin{figure}[t] 
 \centering
 \includegraphics[width=9cm]{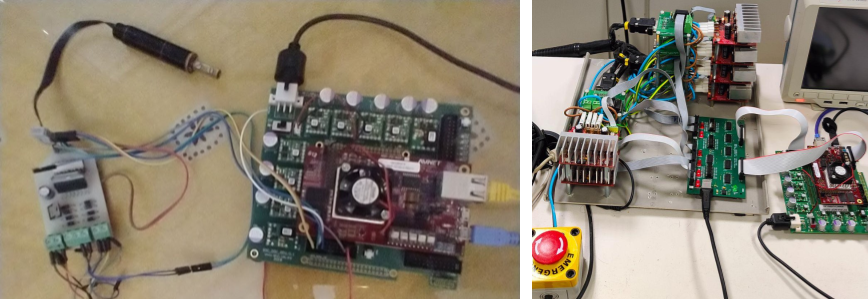}
 \caption{Test setup (left) with a small DC motor and Final setup (right) with Zynq board with carrier board by \href{https://www.t-cober.es}{Cober, S.L.}}
  \label{fig:test}
\end{figure}

\begin{table}
\centering
\caption{Metrics for PL (old and new) and PS (new). }
\begin{tabular}{|c|c|c|c|} 
 \hline
 System & Dynamic (PL) & Dynamic (PS) & Latency \\ 
 \hline
 (New) Zynq & 57mW & 1584mW & 6.5ms\\ 
 (Old) Spartan3+Spartan6 & 157mW  & N/A & 40ms \\ 
 
 \hline
\end{tabular}

\label{t:power_consumption}
\end{table}

A simple test was run using just one spare joint motor to check whether the pipeline worked on the new Zynq board. As we are migrating projects, it is reasonable to expect that, once the new pipeline is working from end to end, the functionality has not been lost. Of course, extensive testing will ensure nothing has gone wrong or is being performed differently, but that will be a matter of tuning and should not affect whether the system works or not.

This test was successful and we were able to command a small DC motor directly from the Zynq board, using a power board as a support for providing the high voltage signals needed to drive the motor. The setup for this test, as well as the motor used, are shown in Figure \ref{fig:test}. Depending on the position commanded, the motor would spin for reaching the commanded angle, meaning the control algorithm was working. 

We have also compared the response speed and estimated power consumption of both old and new infrastructure FPGA wise, not taking into account the high voltage power needed for driving the robot. Results are shown in Table \ref{t:power_consumption}. As can be seen, latency has been significantly reduced, while the PL consumption is significantly less for the new system than for the old one. Furthermore, PS consumption also presents an advantage over a dedicated server consumption, which would have at least a 300W power supply.
\section{Conclusion}

An infrastructure upgrade has been shown in this article, changing the ED-Scorbot's old spiking controller implementation from a system with two Spartan FPGAs, a slow SPI communication system and the necessity of a dedicated server to another with just one Zynq SoC and some support boards, keeping all previous functionalities, mainly the working SPID controller and remote access. AXI replaces the SPI for latency improvement. A successful test has been performed on one small DC motor over different DoF outputs, showing that the new infrastructure is properly working. Overall measured improvements are of 515\% latency in commands reception, and 36\% of the original power consumption for the Programmable Logic.

The Scorbot ER-VII lacks some of the latest upgrades in robotics, such as higher level proprioception. This means that the robot offers a lower lever interface, but we can take advantage of that. Current available robots for purchase for research do not provide the required access to the motors for implementing spiking control to reach motor power with such spikes as muscles do in biological systems. This is one of the main advantages of this neuromorophic robotic platform for the lower power and latency in the control.


However, it could be argued that using a robot this old is counterproductive, given the fact that newly built robots may be cheaper, more efficient and require less infrastructure. Nevertheless, we strongly affirm that this platform can still be used for further research and already has an spiking controller implementation, which is an essential upside. Therefore, we encourage fellow researchers to use the ED-Scorbot platform, as there is only room for improvement and breakthrough. We would like to see this platform as a reference for future work, as it is really groundbreaking and may inspire other spiking controllers, similar or different, that in the end may lead us towards a better understanding of how real biological systems work.



\bibliographystyle{IEEEtran}
\bibliography{IEEEabrv,IEEEexample}
\vspace{12pt}

\end{document}